\documentclass[10pt,twocolumn,letterpaper]{article}

\usepackage{cvpr}
\usepackage{times}
\usepackage{epsfig}
\usepackage{graphicx}
\usepackage{amsmath}
\usepackage{amssymb}
\usepackage[caption=false]{subfig}
\usepackage{algorithm}
\usepackage[noend]{algpseudocode}
\usepackage{array}
\usepackage{ctable}
\usepackage{multirow}
\usepackage[skip=2pt]{caption}
\usepackage[pagebackref=true,breaklinks=true,letterpaper=true,colorlinks,bookmarks=false]{hyperref}
\usepackage[breaklinks=true,letterpaper=true,colorlinks,bookmarks=false]{hyperref}
\usepackage[numbers,sort]{natbib}
\usepackage{balance,flushend}
\usepackage{comment}

\newcolumntype{P}[1]{>{\centering\arraybackslash}p{#1}}

\cvprfinalcopy % *** Uncomment this line for the final submission

\def\cvprPaperID{24} % *** Enter the CVPR Paper ID here

\ifcvprfinal\fi

\begin{document}

\tolerance=999
\sloppy

%%%%%%%%% TITLE
\title{Weakly Supervised Learning Guided by Activation Mapping\\Applied to a Novel Citrus Pest Benchmark}

\author{Edson Bollis\textsuperscript{1} ~~~Helio Pedrini\textsuperscript{2} ~~~Sandra Avila\textsuperscript{1}\\
\textsuperscript{1}REasoning for COmplex Data Lab. (RECOD)  ~~~\textsuperscript{2} Visual Informatics Lab. (LIV)\\
Institute of Computing (IC), University of Campinas (UNICAMP) \\
Campinas, SP, Brazil, 13083-852 \\
\small edsonbollis@gmail.com, \{helio, sandra\}@ic.unicamp.br\\
}

\maketitle
\thispagestyle{empty}

\begin{abstract}
Pests and diseases are relevant factors for production losses in agriculture and, therefore, promote a huge investment in the prevention and detection of its causative agents. In many countries, Integrated Pest Management is the most widely used process to prevent and mitigate the damages caused by pests and diseases in citrus crops. However, its results are credited by humans who visually inspect the orchards in order to identify the disease symptoms, insects and mite pests. In this context, we design a weakly supervised learning process guided by saliency maps to automatically select regions of interest in the images, significantly reducing the annotation task. In addition, we create a large citrus pest benchmark composed of positive samples (six classes of mite species) and negative samples. Experiments conducted on two large datasets demonstrate that our results are very promising for the problem of pest and disease classification in the agriculture field.
\end{abstract}

%%%%%%%%% BODY TEXT
\section{Introduction}

Pests and diseases in orchards are dangerous to the world of agriculture and have caused significant losses. Particularly, the Greening (\textit{Diaphorina citri}), also called Huanglongbing (HLB), --- the actual most destructive disease in citrus agriculture~\citep{GraftonCardwell2018Huanglongbing} --- cost \$13.2 billion to Florida State between 2005 and 2016~\citep{rahmani2017economic}. The real losses are more significant when we consider other pests and diseases that infect the country's production, such as Citrus Variegated Chlorosis (\textit{Xylella fastidiosa}), Citrus Canker (\textit{Xanthomonas axonopodis}), and Citrus Leprosis (\textit{Citrus leprosis virus}).

One way to detect and prevent these threats is the use of Integrated Pest Management (IPM). It describes how to avoid the problems and what are the rules to apply inputs before the problem occurs~\citep{ifas2020Production}. Usually, human inspectors walk along the orchards streets collecting samples to analyze them and reporting the results in paper sheets or mobile tools for data acquisition~\citep{ifas2020Production}. 
The inspectors examine stalks, leaves, and fruits for hours, trying to find mites and insects to quantify them. Depending on the level of the infection, when the number of dispersers (mites or insects) past from a safety limit, the IPM describes the rules to apply inputs, cuts parts of the plant, removes the whole plant or eliminates the plant and its neighborhood. The IPM is a mechanical process that can be done by machines to help small farmers to enforce its rules. In addition, as expected, when humans handle the job, the IPM process is prone to errors due to the inability or fatigue of the handlers.

It is common to see mobile technology in the field to perform a wide range of tasks, such as data acquisition, employee communication, and production management. 
In this scenario, employing mobile devices to detect pests and diseases would not be an additional hurdle. In fact, the use of Convolutional Neural Networks (CNNs) in mobile devices, such as MobileNets~\citep{sandler2018mobilenetv2}, NasNet-A Mobile~\citep{zoph2017learning}, and EfficientNet~\citep{tan2019efficientnet}, can greatly help inspectors in doing their work more efficiently and effectively. 

As a consequence of the lack of other image collections, we created a novel dataset called Citrus Pest Benchmark (CPB). It contains images collected with mobile devices of mites in citrus plants, which is unseen in the literature. Our dataset supports the evaluation of our classification method. Unlike the IP102~\citep{wu2019ip102} database for insect pest recognition, for instance, our benchmark is composed of very tiny regions of interest (mites) compared to the remaining area of the image. In this sense, the straightforward use of CNNs in our citrus pest classification problem would not be efficient. Inspired by recent approaches to cancer classification and object detection~\citep{maicas2019model, pang2019mathcal,sudharshan2019multiple,unel2019power}, we develop a weakly supervised learning method that computes saliency maps to automatically locate patches of interest in the original images.

The main contributions of our work are: (i) creation of a new benchmark for the citrus pest recognition problem, where tiny regions of interest containing different types of mites are present in the original images; (ii) development of a weakly supervised multiple instance learning method guided by saliency maps to automatically identify patches in the images and reduce the task of image labeling; (iii) implementation of a weighted evaluation strategy for properly generating a final probability for every extracted image patch; and (iv) achievement of promising classification results on two large pest benchmarks in the agriculture field.

This text is organized as follows. In Section~\ref{sec:related_work}, we briefly review some relevant concepts and approaches related to disease and pest classification and multiple instance learning. In Section~\ref{sec:our-database}, we describe our Citrus Pest Benchmark. In Section~\ref{sec:our-method}, we present our weakly supervised multiple instance learning method. We report and discuss the experimental results achieved on two datasets in Section~\ref{sec:results}. Finally, some concluding remarks and directions for future work are presented in Section~\ref{sec:conclusions}.

\section{Related Work}
\label{sec:related_work}

In this section, we first overview the literature of disease and pest classification, in particular we focus on CNN-based approaches. Then, we describe relevant aspects related to multiple instance learning and weakly supervised approaches.

\subsection{Disease and Pest Classification}

In the era of Convolutional Neural Networks (CNNs), the first works on disease and pest classifiers have the primary goal of improving the classification metrics on a given database. As CNNs require a large amount of training data, many approaches have focused their efforts on creating image databases for classifying pests and diseases in the field (for instance, \citep{hughes2015AnOpen,alfarisy2018deep,wu2019ip102}). 

Concerning disease classification, \citet{hughes2015AnOpen} created an image database called PlantVillage, which consists of 55,000 images (captured in laboratories) from disease symptoms in leaves. \citet{mohanty2016UsingDeep} used the Inception~\citep{szegedy2015going} and AlexNet~\citep{krizhevsky2012imagenet} networks to train their models on the PlantVillage. \citet{ferentinos2018deep} introduced a new version of the PlantVillage with 87,848 images (not publicly available) to evaluate CNNs for plant disease detection and diagnosis. They also proposed its use in mobile applications, but they did not present any experiments. The PlantVillage database was the first large-public database on disease detection area, and many works evaluated well-known CNNs with little or no modification~\citep{Nachtigall2017ClassificationofApple,liu2017identification,Tan2016IntelligentAlerting,ma2018recognition,bhandari2017TowardsCollaborationBetween}. 

With respect to pest classification, before 2018 few works explored CNNs. \citet{liu2016localization} used saliency maps constructed by a histogram. They used the color variation between the pests and backgrounds to extract paddy pests and created a database of 5,136 images. \citet{alfarisy2018deep} collected from Internet 4,511 images of paddy pests and classified them with a CNN. 

\citet{lee2018study} created a pest tangerine database of 10 macro-insects, including the Psyllid (\textit{Diaphorina Citri}, the greening vector) and they evaluated several CNNs on their data. Similar to ours, \citet{li2019coarse} proposed a database in which the insects are very tiny concerning to the entire image. They applied a two-stage object detector to find groups of insects in the images and then extracted these regions to detect each insect. In contrast to our approach, we do not have object annotations, so we benefit from a weakly supervised method to classify the pests. \citet{chen2020agricultural} used the Google image search engine to collect 700 images from four pests, including Spider mites (\textit{Tetranychidae}). They used CNNs to classify the images captured from sensors in the field, but they did not show any results related to these types of images.

The largest database for insect pest classification was introduced by~\citet{wu2019ip102}. The IP102 database consists of 102 classes and 75,222 images. The authors applied different CNNs (AlexNet, GoogleNet, VGGNet, and ResNet) to report their results. \citet{ren2019feature} improved the classification performance on IP102 by modifying ResNet blocks. \citet{xu2019xcloud} used the IP102 dataset to demonstrate the use of XCloud, a cloud platform proposed to facilitate the use of AI.

In brief, agricultural works on the Machine Learning area lack of proposition on new methods. Usually, the works only apply the well-known CNNs in its databases. To the best of our knowledge, no work uses mite images collected with mobile cameras using a strict protocol directly in the field.

\subsection{Multiple Instance Learning-based Approaches}
\label{sec:mil-related-work}

Multiple instance learning (MIL) is a weakly supervised category of problems where its training data is arranged in \emph{bag sets} and sets of patches from the bags, called \emph{instances}. The labels are provided only for the bags and the instances inherit from the bags creating a weakly supervised environment~\cite{carbonneau2018multiple}. 

The standard MIL assumption, in a binary problem, states that negative bags contain only negative instances and positive bags contain at least one positive instance. This assumption can be relaxed to use the evaluation of the interaction of several positive instances, as we use in this work~\cite{foulds2010review}.

\citet{sun2016multiple} proposed a weakly supervised CNN framework, called Multiple Instance Learning Convolutional Neural Networks (MILCNN), which fuses residual network and multiple instances learning loss layer. The architecture received the number of instances from a bag, inferring the instances as separated images, and used a function to mix the probabilities to calculate a final probability for the entire bag in the last layer.

\citet{choukroun2017mammogram} introduced an MIL method for mammogram classification using a VGGNet followed by a refining fully connected neural network modified to the MIL paradigm.

\citet{he2019midcn} created a Multiple Instance Deep Convolutional Network for image classification (MIDCN) based on a feature extractor from CaffeNet. They calculated the differences between feature vectors from instances and pre-calculated features, called prototypes, and predicted the classes using these differences. \citet{li2019deep} developed an attention-based CNN model for MIL, which used an adaptive attention mechanism into a CNN to detect significant instances for histopathology images.

The previously mentioned works used all instances of one bag at the same time in the training phase, as a batch of instances. For this, the researchers must adapt the original CNNs changing the first layers and the loss functions. In our proposal, we use the CNN architecture in its original version.

Out of the MIL methods, some works used the same idea of patches, however, in supervised ways as~\cite{pang2019mathcal,unel2019power}.
It was not different for disease and pest classification, as~\citet{li2019coarse}. Unlike the other, \citet{liu2016localization} used a weakly supervised method based on saliency maps to cut the pest from the original images to create their dataset.

According to~\citet{zhou2018semi}, we can classify our transfer learning technique as a pseudo-label for CNNs. However, most of the pseudo-label works came from the inference of the unlabeled part of the databases for models already trained with the labeled part, for instance, the approach developed by~\citet{lee2013pseudo}. In the case of our work, we use pseudo-labels from the original bags. \citet{tao2018zero} and~\citet{zhang2020progressive} also used pseudo-labels with multiple instances in their projects, but not as our work does. 

To the best of our knowledge, we have found neither MIL methods applied to disease and pest classification tasks nor MIL architectures for mobile devices, which encourages our investigation into these research topics.

\section{Our Citrus Pest Benchmark}
\label{sec:our-database}

\begin{figure}[!htb]
\captionsetup[subfloat]{farskip=2pt,captionskip=2pt}
\centering
\subfloat[Magnificent glass]{\includegraphics[height=2.8cm,width=3.8cm]{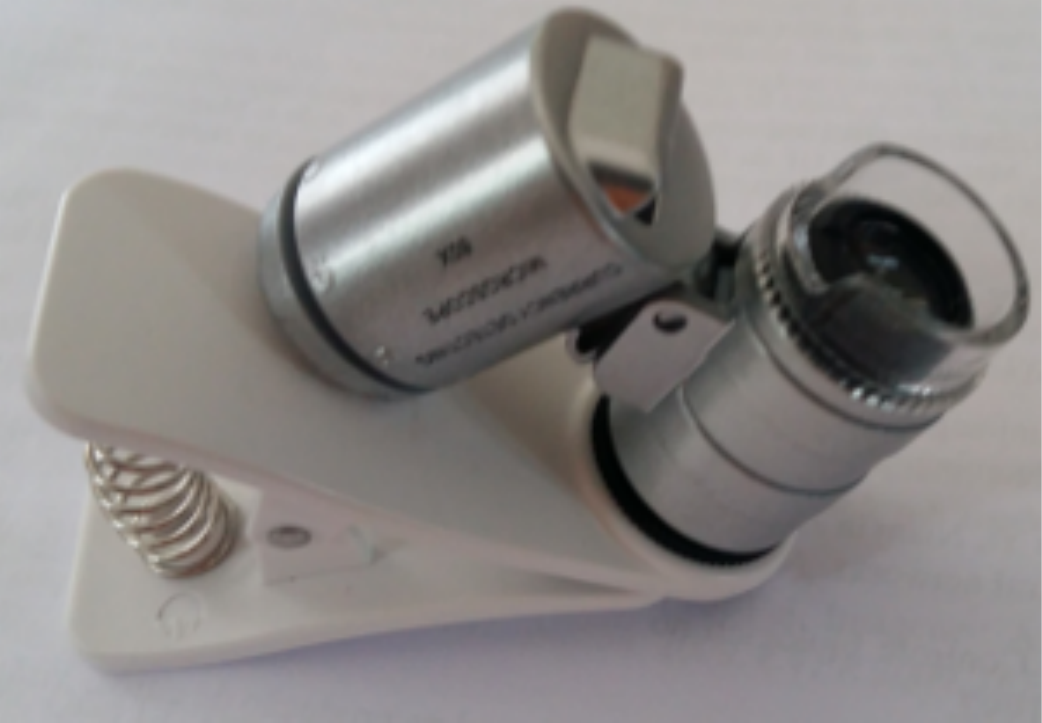} \label{pics:lupa-a}} \hspace*{0.1cm}
\subfloat[Mobile coupled]{\includegraphics[height=2.8cm,width=3.8cm]{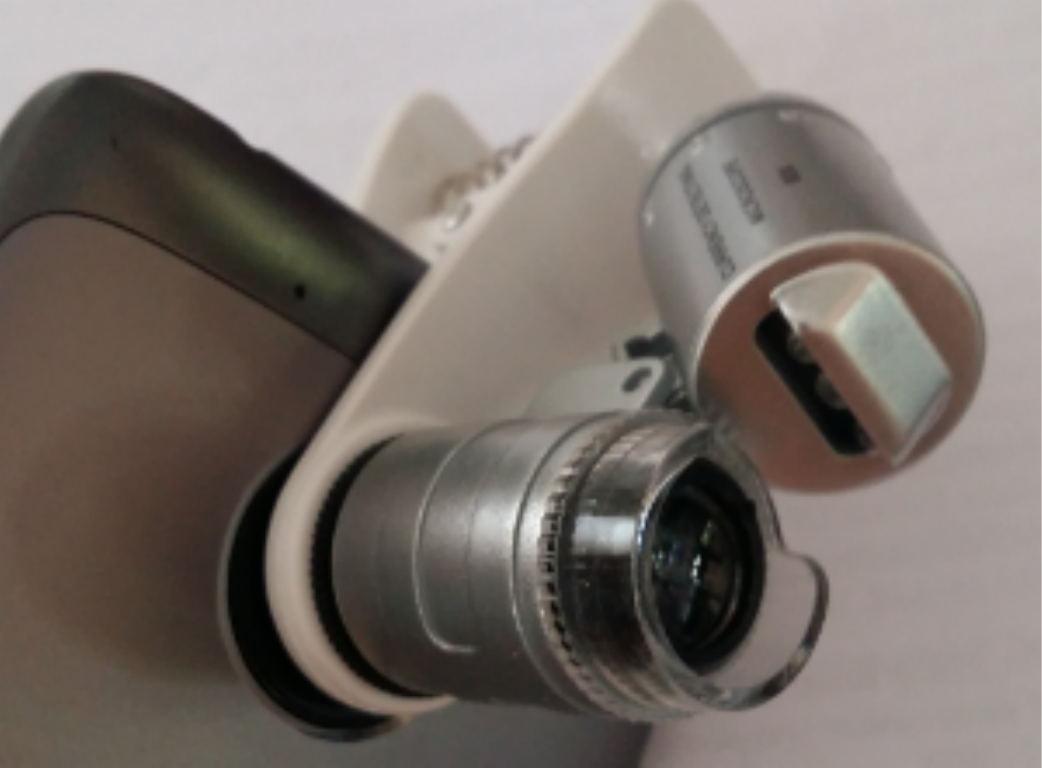}\label{pics:lupa-b}} \\
\subfloat[Insect with normal size]{\includegraphics[height=2.8cm,width=3.8cm]{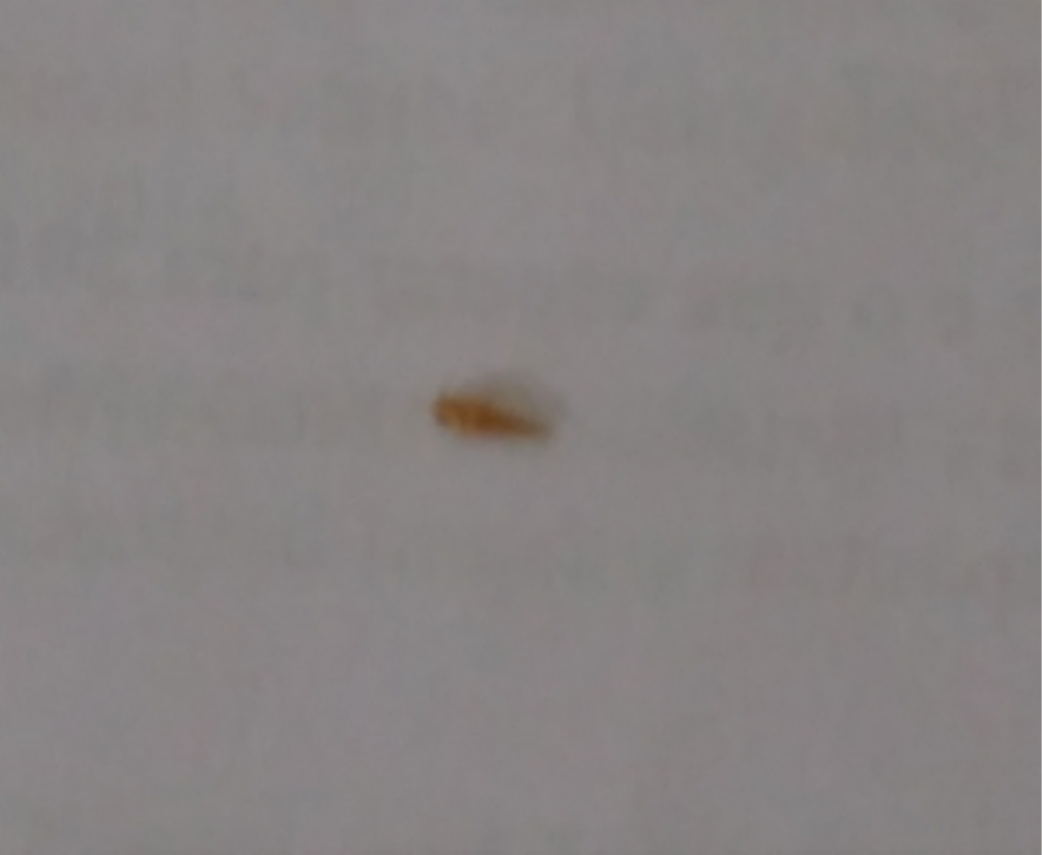}\label{pics:lupa-c}} \hspace*{0.1cm}
\subfloat[Insect after zoom (60$\times$)]{\includegraphics[height=2.8cm,width=3.8cm]{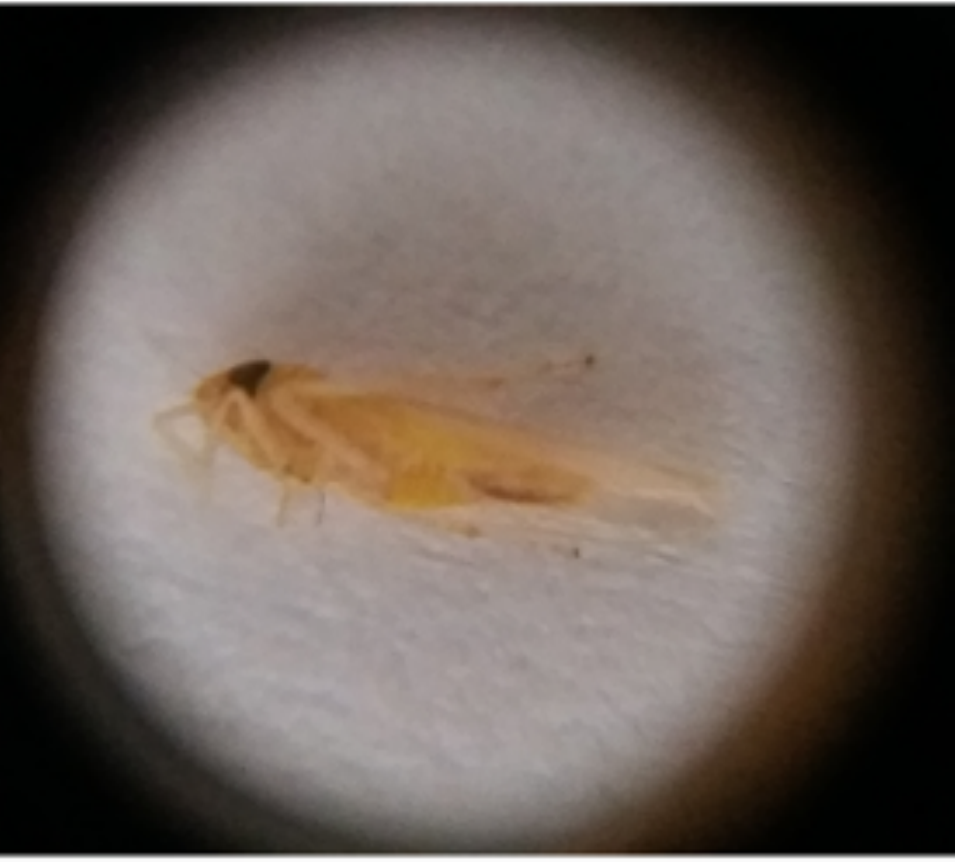} \label{pics:lupa-d}}
\caption{(a-b) Devices used to collect the citrus pest images; (c-d) insect image before and after magnification.}
\label{pics:magnificent_glass}
\end{figure}

As an additional contribution of this work, we created a benchmark\footnote{\url{https://github.com/edsonbollis/Citrus-Pest-Benchmark}.} containing images divided into seven classes (six mite species and a negative class). The images were collected via a mobile device coupled with a lens magnifier, as shown in Figure~\ref{pics:magnificent_glass}. 
In the acquisition process, we employ a Samsung Galaxy A5 with a 13~MP camera coupled with a 60$\times$ magnifier, equipped with a white LED lighting and ultraviolet LED.

\begin{figure}[!htb]
\captionsetup[subfloat]{farskip=2pt,captionskip=2pt}
\centering
\subfloat[Red Spider]{\includegraphics[height=2.8cm,width=2.6cm]{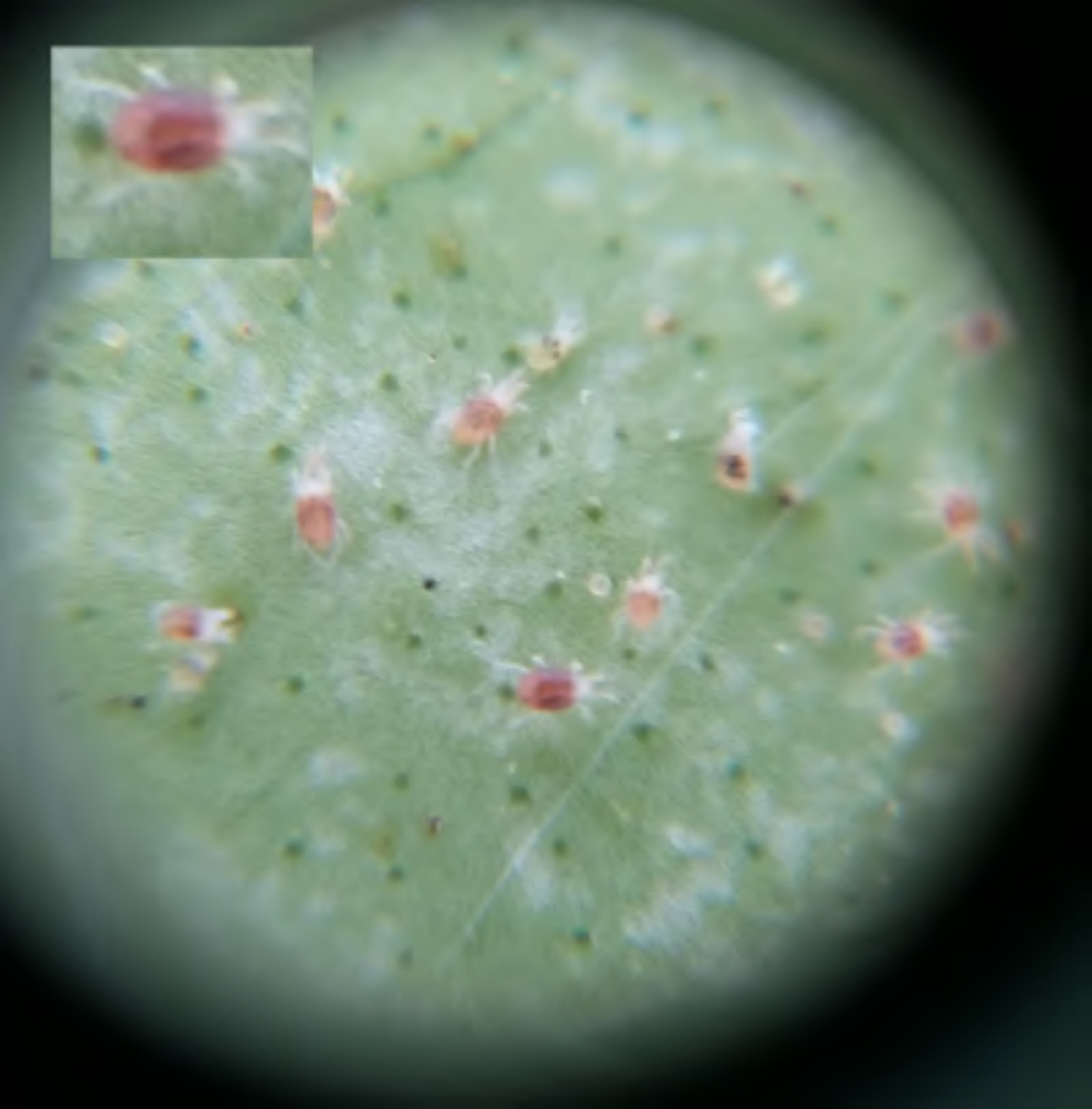}\label{pics:classes-a}} \hspace*{0.1cm}
\subfloat[Phytoseiid]{\includegraphics[height=2.8cm,width=2.6cm]{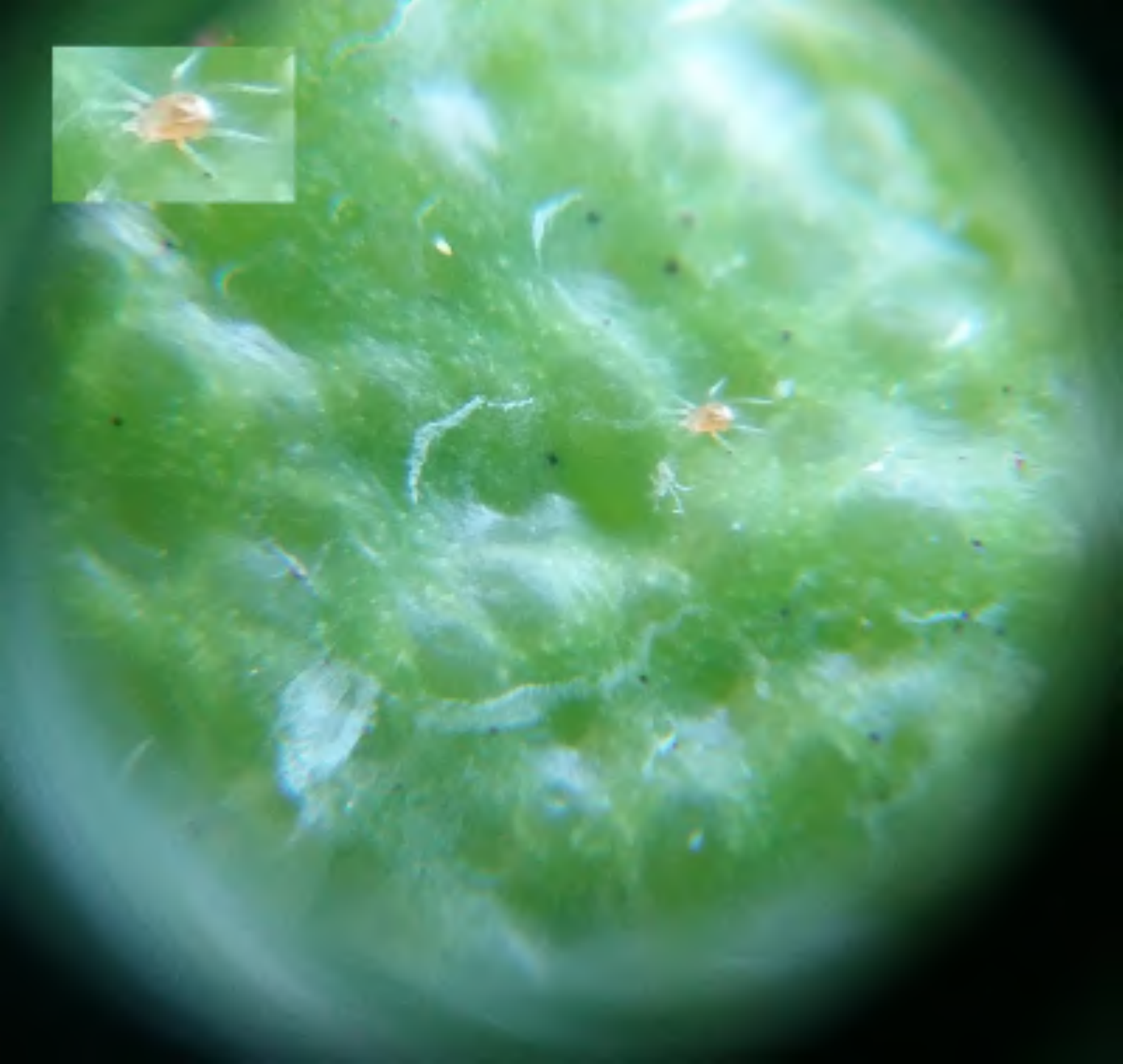}\label{pics:classes-b}} \hspace*{0.1cm}
\subfloat[Rust]{\includegraphics[height=2.8cm,width=2.6cm]{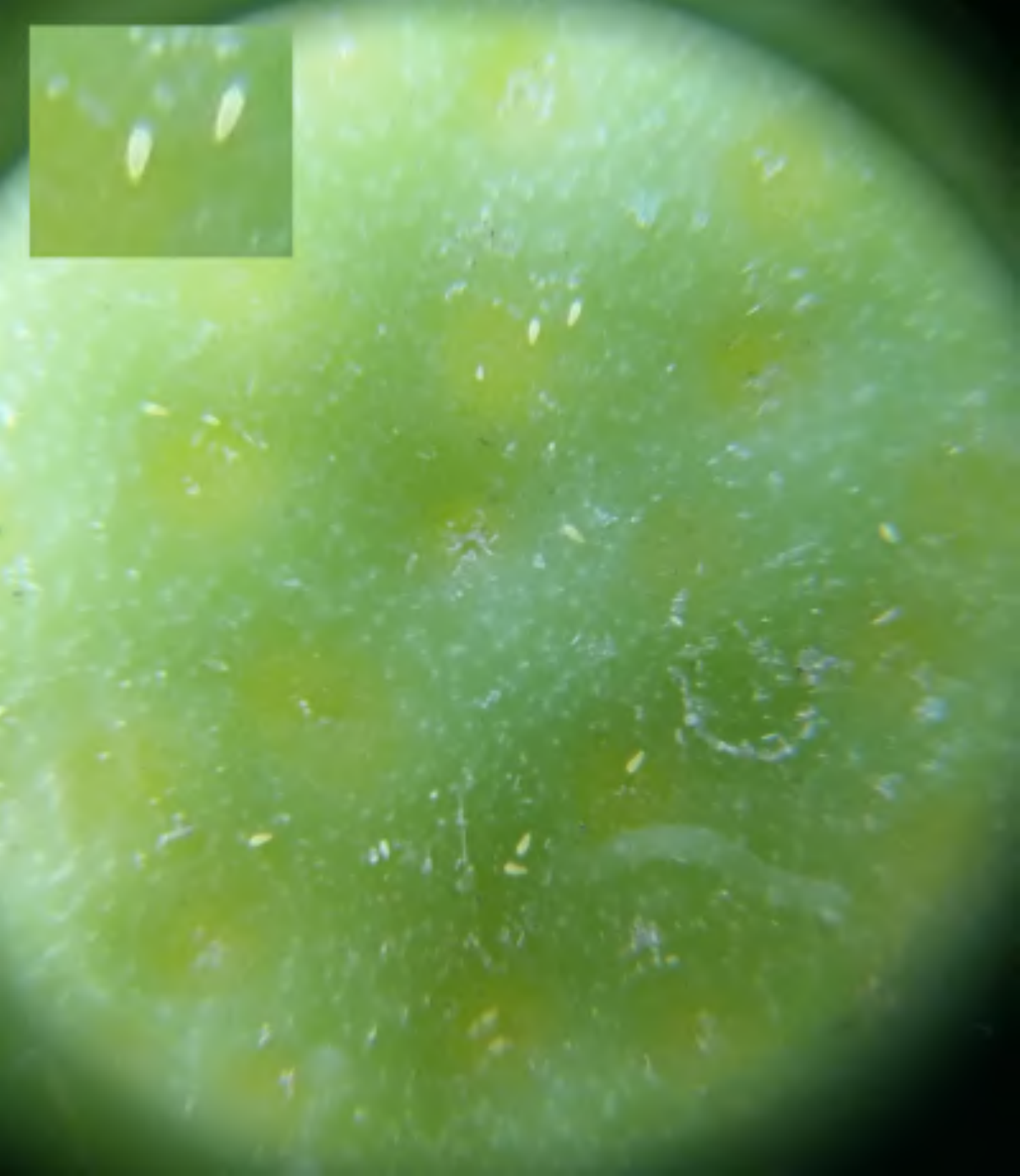}\label{pics:classes-c}} \\
\subfloat[False Spider]{\includegraphics[height=2.8cm,width=2.6cm]{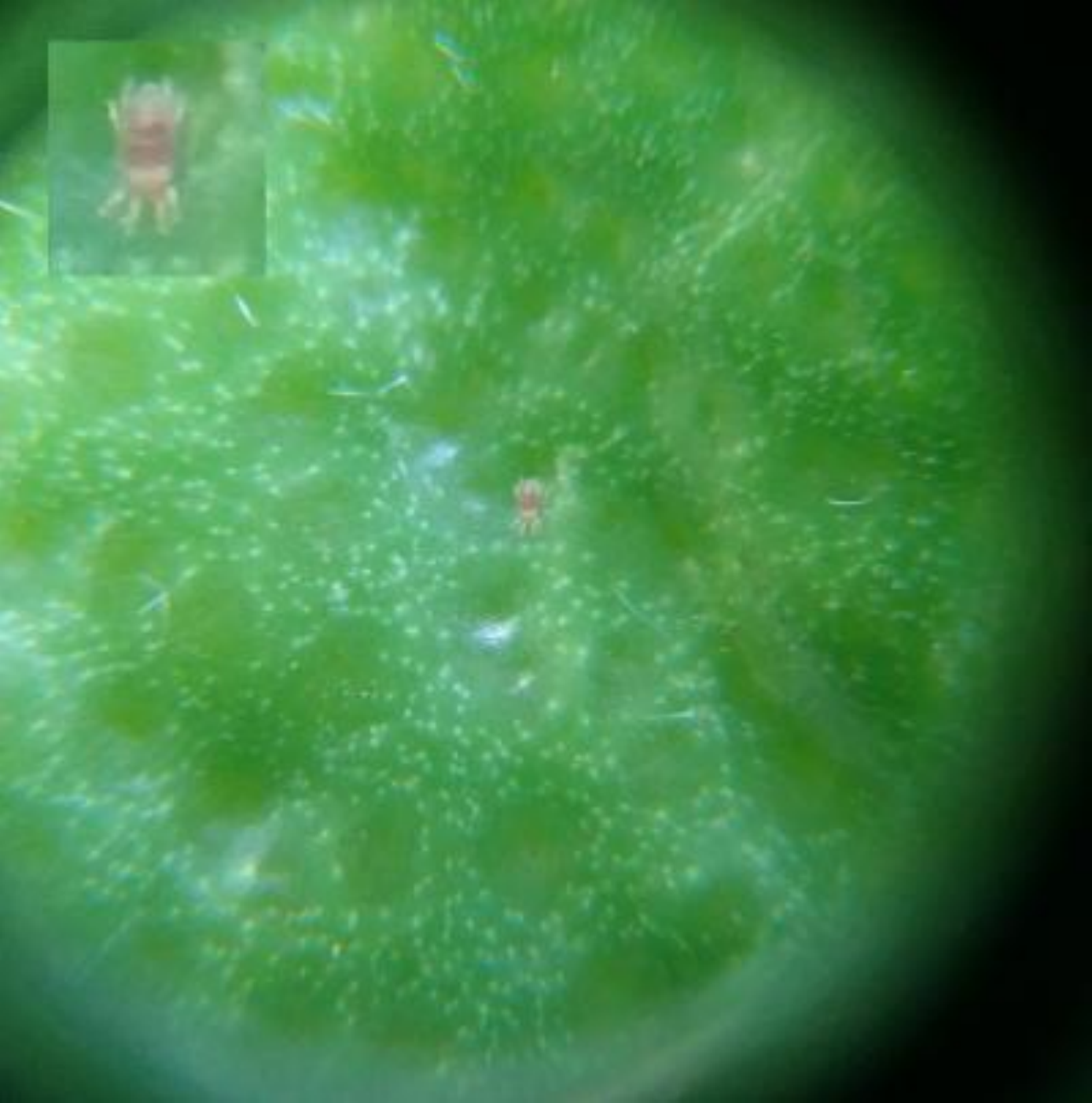}\label{pics:classes-d}} \hspace*{0.1cm}
\subfloat[Broad]{\includegraphics[height=2.8cm,width=2.6cm]{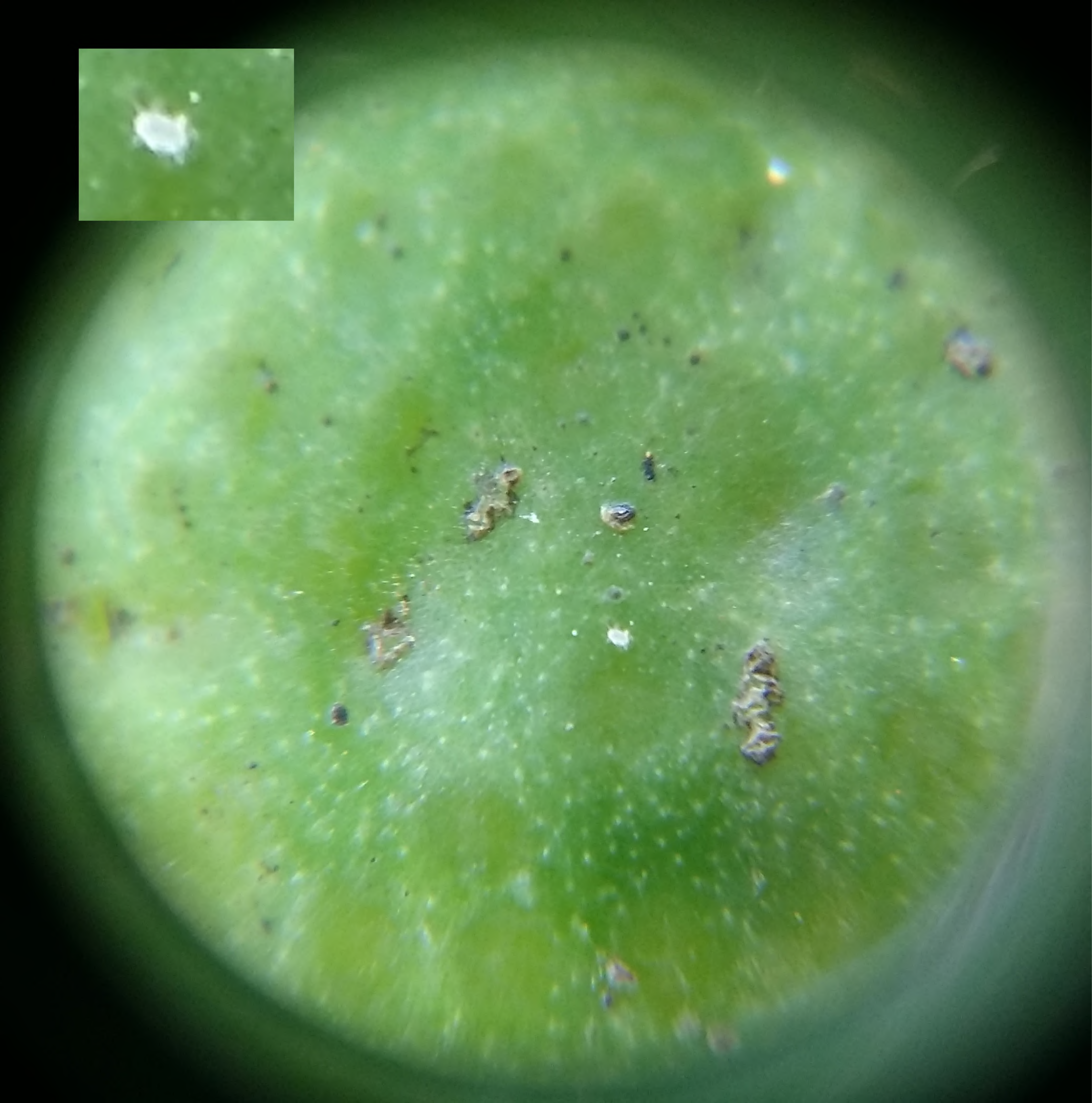}\label{pics:classes-e}} \hspace*{0.1cm}
\subfloat[Two-Spotted Spider]{\includegraphics[height=2.8cm,width=2.6cm]{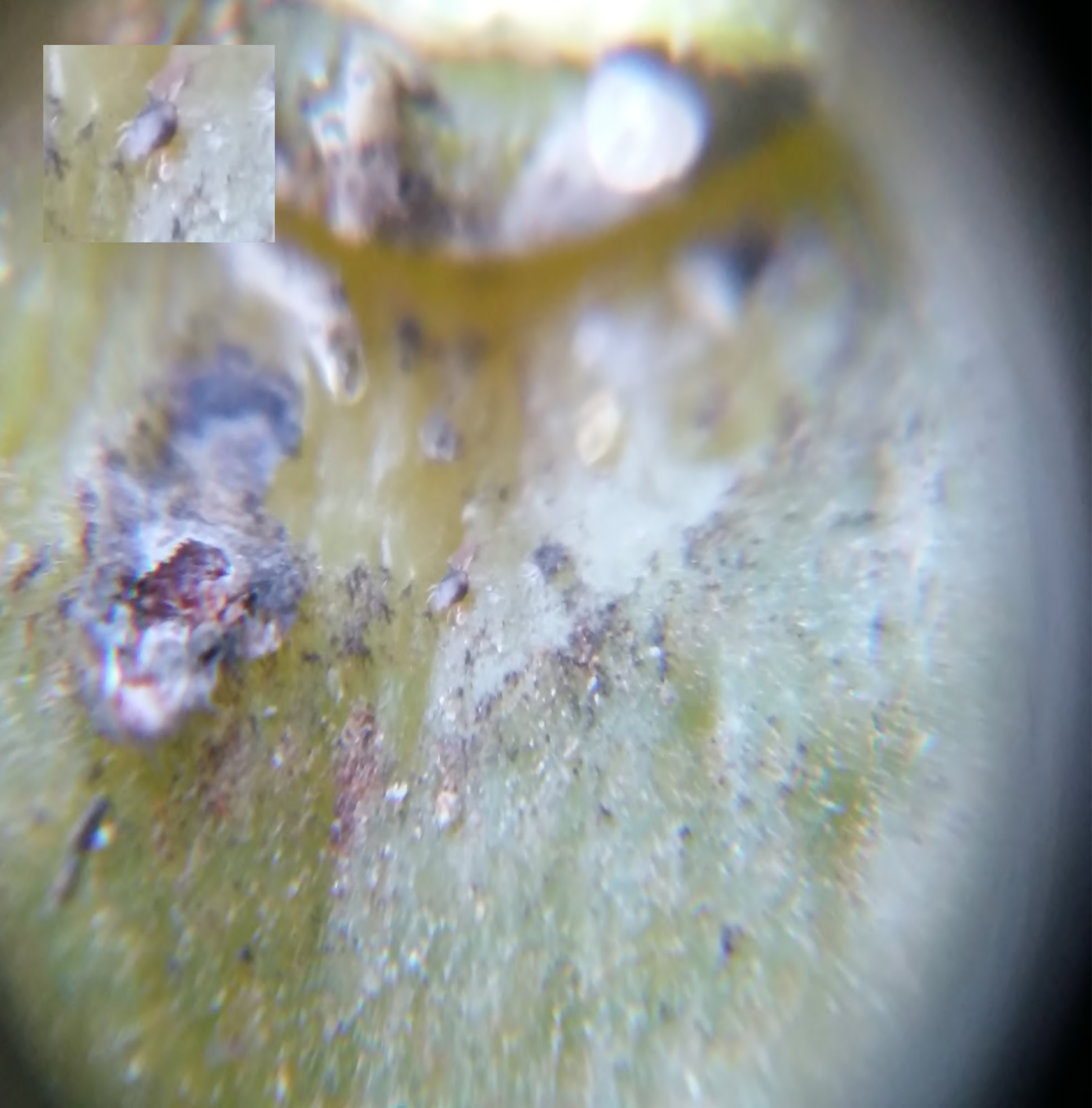}\label{pics:classes-f}}
\caption{Mites captured through optical magnification of 60$\times$. The mites are highlighted on the upper-left side of the images.}
\label{pics:classes}
\end{figure}

\begin{figure}[!htb]
\captionsetup[subfloat]{farskip=2pt,captionskip=2pt}
\centering
\subfloat[Sharp image]{\includegraphics[height=2.8cm,width=3.8cm]{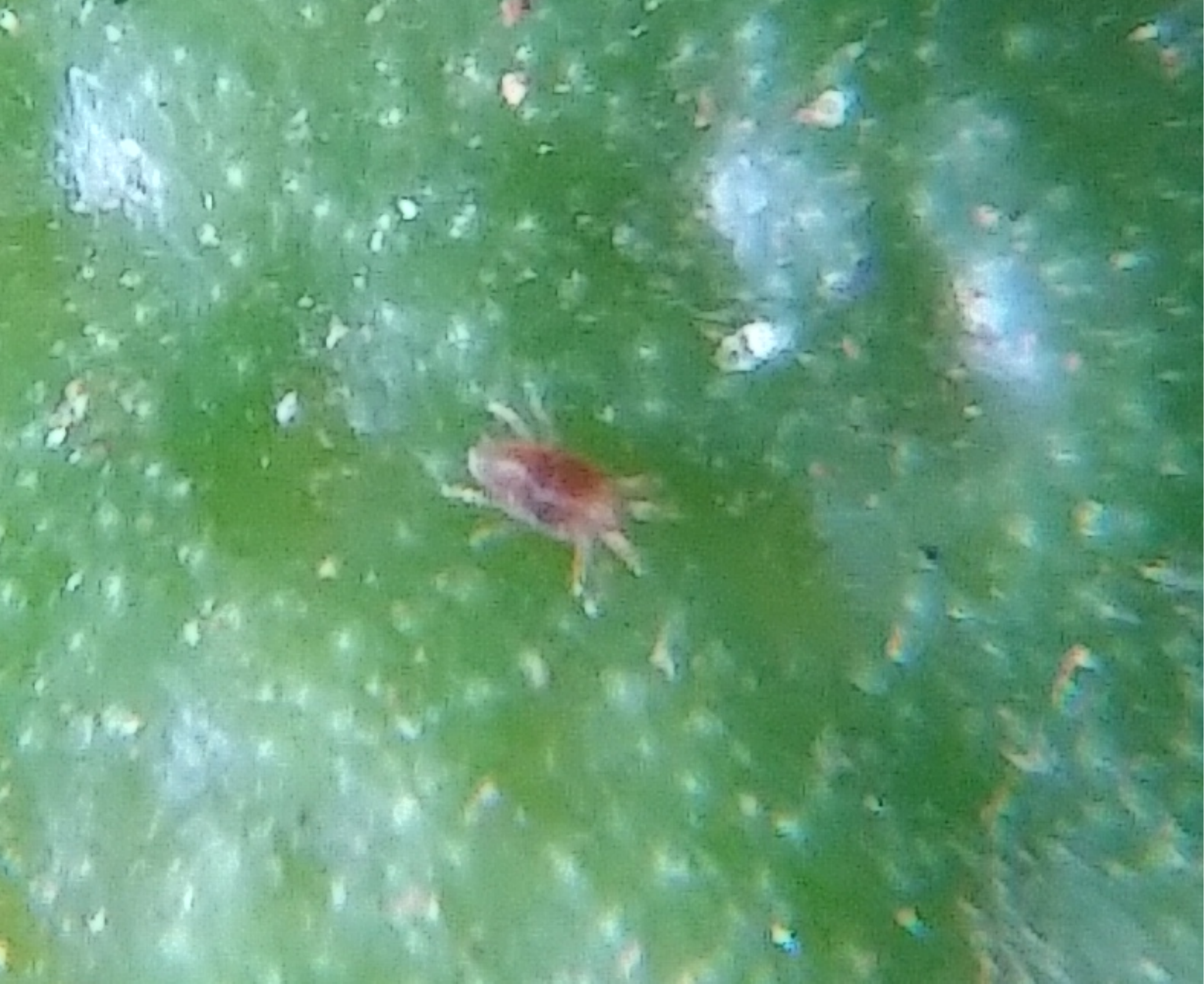}\label{pics:good_bad_image-a}} \hspace*{0.1cm}
\subfloat[Blurred image]{\includegraphics[height=2.8cm,width=3.8cm]{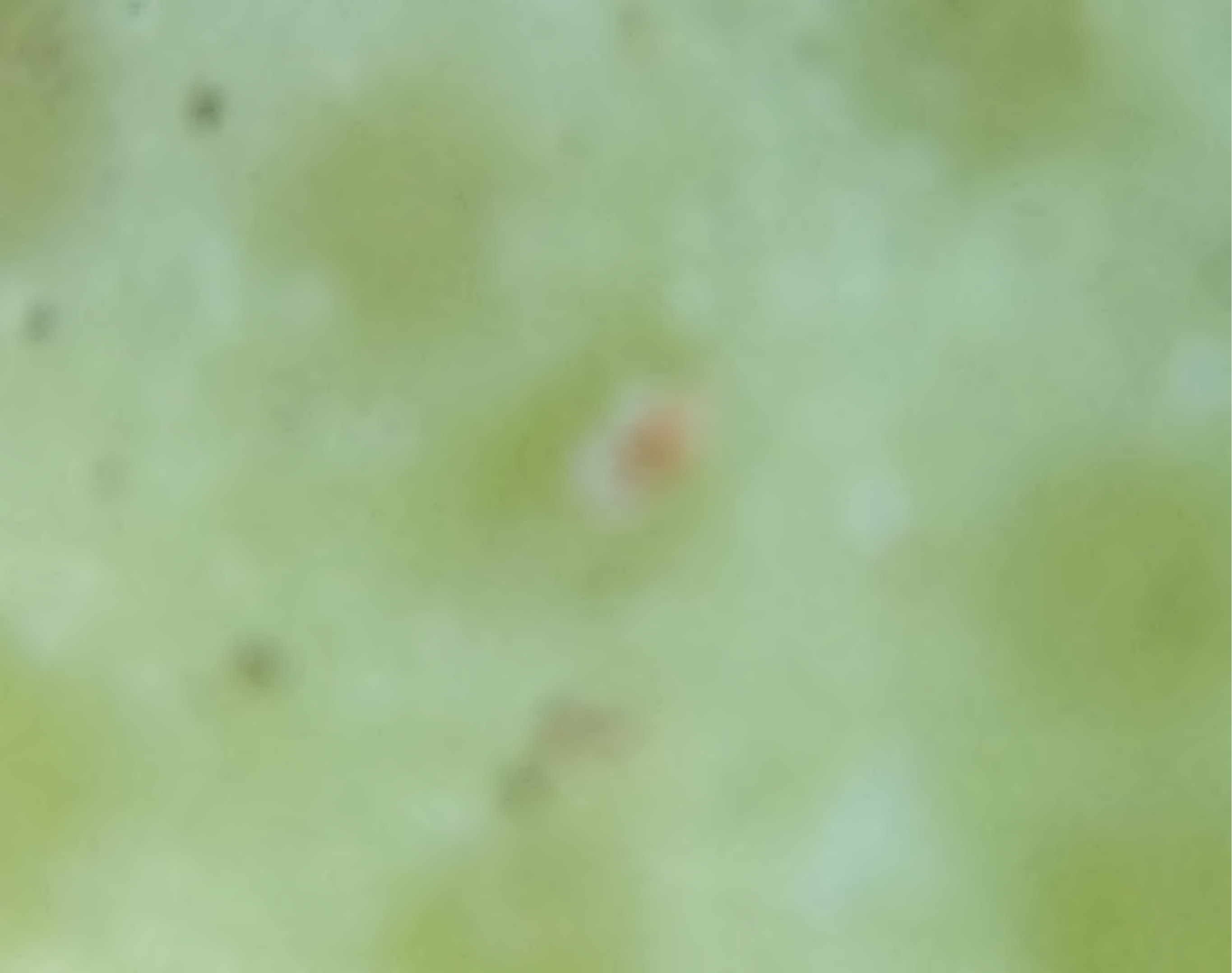}\label{pics:good_bad_image-b}}
\caption{Samples of False Spider mites from our Citrus Pest Benchmark.}
\label{pics:good_bad_image}
\end{figure}

The sizes of the mite species are very small in proportion to the entire image size, as illustrated in Figure~\ref{pics:classes}. Due to the hard glass surface present in the device, a significant part of the images is blurred, as can be seen in Figure~\ref{pics:good_bad_image}.

To generate our citrus pest database, the mite images were captured at S\~ao Jos\'e Farm, located in the city of Rio Claro, S\~ao Paulo State, in Brazil. The data acquisition period was from March~2018 to January~2019. Guided by MIP inspectors, we carried out scheduled inspections in the production unit areas, which contain up to 1000 citrus trees divided into groups arranged in lines. The inspectors chose samples from the crop lines, not near the border, to analyze the fruits, new germinations and stem. Then, they moved on to the next thirtieth plant individuals. We used the samples examined by the inspectors to obtain the mite images. After completing a crop sector line, every three planting lines were examined.

Our database consists of 10,816 multi-class images categorized into seven classes: (i) 1,902 images with Red Spider mites (\textit{Panonychus citri}, \textit{Eutetranychus banksi}, \textit{Tetranychus mexicanus}), the largest of all other species which produces a yellowish symptom on the leaves and fruits (Figure~\ref{pics:classes-a}); (ii) 1,426 images with Phytoseiid mites (\textit{Euseius citrifolius}, \textit{Iphiseiodes zuluagai}), the predator mite that helps control other mites (Figure~\ref{pics:classes-b}); (iii) 1,386 images with Rust mites (\textit{Phyllocoptruta oleivora}), responsible for the rust symptom and significant crop losses (Figure~\ref{pics:classes-c}); (iv) 1,750 images with False Spider mites (\textit{Brevipalpus phoenicis}), a vector of the Leprosis virus (Figure~\ref{pics:classes-d}); (v) 806 images with Broad mites (\textit{Polyphagotarsonemus latus}), responsible for causing a white cap on the fruits (Figure~\ref{pics:classes-e}); (vi) 696 images with Two-Spotted Spider mites, which do not bring significant crop losses, however, they are clearly visible in the field (Figure~\ref{pics:classes-f}); and (vii) 3,455 negative images.

We partitioned the image collection into three groups, referred to as training, validation and test, containing approximately 60\%, 20\%, and 20\% of the mites from each class totaling 6380, 2239 and 2197 images, respectively.

Some of the classes are very similar to each other for untrained eyes. In addition, the differences in luminosity and zoom make the database very challenging. The multi-class problem turns the tasks more interesting once we have 5\% (599) of images with up to three classes simultaneously.

Although we collected the images with the aid of human inspectors, the errors inter-classes are significant due to the size of the mites. The inspectors are currently revising the multi-class labels and, for this reason, we are publishing images of 1,200$\times$1,200 pixels for the negative and positive classes, more precisely, 7,966 mite images and 3,455 negative images.

In Table~\ref{table:databases}, we compare our benchmark to various existing databases related to the pest and disease recognition task and cited in our work.

\begin{table*}[!htb]
\setlength{\tabcolsep}{3.0mm}
\centering
%\small
\begin{tabular}{lcrcc}
\toprule
Author & Database Name & Size & Type & Year \\
\midrule
\citet{hughes2015AnOpen} & PlantVillage & 55,000 & symptoms of diseases & 2015 \\
\citet{Barbedo2016IdentifyingMultiple} & N/A & 1,335 & symptoms of diseases & 2016 \\
\citet{Nachtigall2017ClassificationofApple} & N/A & 2,539 & symptoms of diseases & 2016 \\
\citet{Tan2016IntelligentAlerting} & N/A & 4,000 & symptoms of diseases & 2016 \\
\citet{liu2016localization} & Pests ID & 5.136 & pests & 2016 \\
\citet{bhandari2017TowardsCollaborationBetween} & N/A &  \multicolumn{1}{c}{N/A} & symptoms of diseases & 2017 \\
\citet{liu2017identification} & N/A & 13.689 & symptoms of diseases & 2018 \\
\citet{alfarisy2018deep} & Paddy Pest Image & 4,511 & pests & 2018 \\
\citet{lee2018study} & Pest Tangerine & 5,247 & pests & 2018 \\
\citet{wu2019ip102} & IP102 & 75,222 & pests & 2019 \\
\citet{li2019coarse} & Aphid Images & 2,200 & pests & 2019 \\
\citet{chen2020agricultural} & N/A & 700 & pests & 2020 \\
Our work & CPB & 10,816 & pests & 2020 \\
\bottomrule
\end{tabular}
\caption{Pest and disease databases. N/A means that the value was not available from the original paper.}
\label{table:databases}
\end{table*}

\section{Methodology}
\label{sec:our-method}

\begin{figure*}[!htb]
\centering
\includegraphics[width=1.00\textwidth]{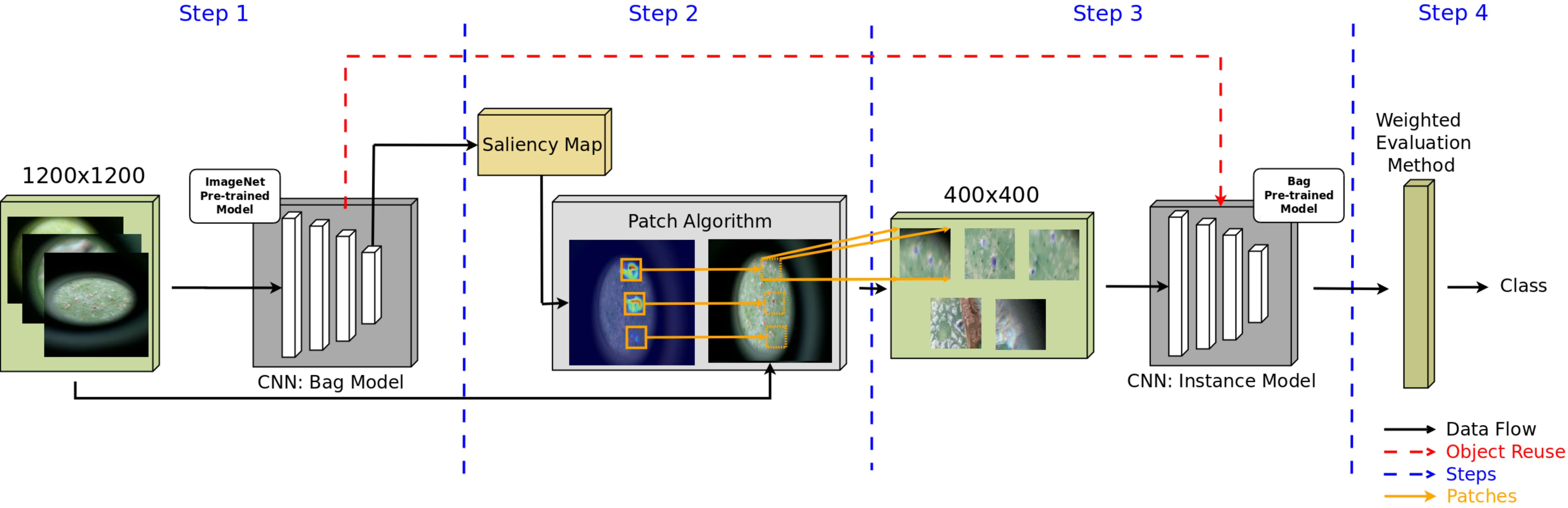}
\label{pics:transf-a}\hspace{0cm}
\caption{Our method consists of four steps. In Step 1, we train a CNN (initially trained on the ImageNet) on the Citrus Pest Benchmark. In Step 2, we automatically generate multiple patches regarding saliency maps. In Step 3, we fine-tune our CNN model (trained on the target task) according to a multiple instance learning approach. In Step 4, we apply a weighted evaluation scheme to predict the image class.}
\label{pics:pipeline}
\end{figure*}

In this section, we introduce our weakly supervised approach, which is guided by saliency maps. In Section~\ref{sec:high-algorithm}, we describe our problem within the framework of multiple instance learning (MIL). Next, in Section~\ref{sec:salimap_cut}, we detail the proposed Patch-SaliMap, a multi-patch selection strategy based on saliency maps. Finally, in~Section~\ref{sec:hierar-eval}, we~explain how to evaluate an image considering the generated~patches. We depict the main stages of our pipeline in Figure~\ref{pics:pipeline}.

\subsection{Multiple Instance Learning Framework}
\label{sec:high-algorithm}

In brief, our method consists of four steps: (1) we train a CNN (initially trained on the ImageNet) on the Citrus Pest Benchmark, (2) we automatically generate multiple patches regarding saliency maps, (3) we fine-tune our CNN model (trained on the target task) according to a multiple instance learning approach, and (4) we apply a weighted evaluation scheme to predict the image class.

As mentioned before, multiple instance learning (MIL) is a form of weakly supervised learning where training data is a set of labeled \emph{bags} $X = \{x_{i}, i=1,...,n\}$, and each bag contains several \emph{instances} $\overline{X} = \{x_{ij}, j = 1,...,k\}$, where $x_{ij}$ is part of $x_{i}$, $n$ is the number of images, $k$ is the number of instances, and $j$ is the number of images from $\overline{X}$. In this context, in Step 1, the CNN model (trained on a set of labeled bags) is our Bag Model. 

In Step 2, we generate patches from the bags, as detailed in Section~\ref{sec:salimap_cut}. Our algorithm uses the saliency of the maps to identify the regions on the images where mites are highly likely to be located. In other words, we apply the algorithm in each $x_i \in X$ to generate $\{x_{ij}, j=1,...,k\}$ patches of~$x_i$, with $k = 5$. Thus, we create a new instance database $\overline{X} = \{x_{ij}, i =1,...,n, j=1,...,k\}$ for MIL.

In Step 3, we assume the class label of an instance is the same of its bag (in MIL the labels are only assigned to bags). That is, if $y_i = f(x_i)$ is the label of $x_i \in X$ then $f(x_{ij}) = y_i,~x_{ij} \in \overline{X}$. Next, we finetune the same Bag Model on the $\overline{X}$, exploring a transfer learning scheme to MIL. 
Since we have more mites than negative images, we use five instances of each negative bag and two instances of positive bags to balance the data and decrease the probability to miss mites on positive images.

We highlight that it is possible to use the same model with no changes for images with different sizes because there is a global pooling after the last convolutional layer for every CNN. The pooling transforms a feature map of dimension $w \times h \times c$ in a feature map of size $1 \times 1 \times c$. Therefore, we are able to reuse bag models and instances regardless of the image sizes.

In Step 4, all the models trained in $\overline{X}$ are evaluated on its subsets that contain patches of $X$, producing the evaluation for the bags, as described in Section~\ref{sec:hierar-eval}. The best model evaluated in $\overline{X}$ is referred to as Instance Model and it provides a final probability for every instance and, applying the proposed Weighted Evaluation Method, for every bag.

\subsection{Multi-patch Selection Strategy Based on Saliency Maps}
\label{sec:salimap_cut}

Our aim here is to learn fine-grained details since most citrus mites are not readily visible to the naked eye. We propose to select significant image patches according to the saliency map, called the \emph{Patch-SaliMap} algorithm. In Algorithm~\ref{alg:Patch-SaliMap}, we formally describe our proposal.

Let $x_i \in X \subset \mathbb{R}^{h\times w \times 3}$ be a tensor of an image, where $h,w \in \mathbb{N}^+$ are the height and width of $x_i$. 
Let $S: \mathbb{R}^{h\times w \times 3} \rightarrow \mathbb{R}^{h\times w}$ be a saliency map function, where $S(x_i)$ is the saliency map of $x_i$. 
The \textit{Patch-SaliMap} takes as input $x_i, S(x_i), k, l$ and produces $\{x_{ij} \in \overline{X} \subset \mathbb{R}^{l\times l \times 3}, j = 1,...,k\}$, where
$k \in \mathbb{N}^+$ is the total number of instances,
$j \in \mathbb{N}^+$ is the index of instances, and 
$l \in \mathbb{N}^+$ is the height of a square patch.

The algorithm uses the prior knowledge that the mites are small enough to fit in images with a size smaller than the patch size, $l \times l$ pixels. As a consequence, the algorithm achieves a higher probability of obtaining instances with mites in the first patches. 

Using the maximum of the saliency map matrix is an excellent choice at the inference time. However, when we are training the Instance Model, the regions of the maximum gradient for negative instances usually bring features easy to learn, which makes the algorithm addicted to find these features only. Thus, to fix it for the training set of $\overline{X}$, we produce random patches $x_{ij}$, where $x_i$ has negative labels to force the Instance Model to learn more robust features.

\begin{algorithm}
\small
\caption{Patch-SaliMap}\label{alg:Patch-SaliMap}
\hspace*{\algorithmicindent} \textbf{Input:} $x_i, S(x_i), k, l$\\
\hspace*{\algorithmicindent} \textbf{Output:} instances
\begin{algorithmic}[1]
\Function{Patch\_SaliMap:}{}
\State $l = \textit{l} / 2$
\For{$i := 1:k$} 
\State $a,b :=\textit{ get maximum indices from values of }S(x_i)$
\If {$a \pm l,b \pm l \textit{ is out of } x_i \textit{ border}$}
\State $a,b := \textit{fix a,b using l}$
\EndIf
\State \# get a new patch around the indices
\State $\textit{new patch} := x_i[a-l:a+l,b-l:b+l,:]$
\State $\textit{min} := \textit{get minimum value of } S(x_i)$
\State \# occlude using saliences
\State $S(x_i)[a-l:a+l,b-l:b+l,:] := \textit{min}$
\State $\textit{instances}[i] := \textit{new patch}$
\EndFor
\\
\Return \textit{instances}
\EndFunction
\end{algorithmic}
\end{algorithm}

\subsection{Weighted Evaluation Method} \label{sec:hierar-eval}

To predict the class of bag images, we propose the Weighted Evaluation method. It uses static weights to calculate a weighted average and reports the final probabilities. Thus, given $x_{i} \in X, i=1,...,n$, its $x_{ij} \in \overline{X}, j=\{1,...,k\}$, and the probabilities $p(.)$ from the Instance Model, the final probability $P(.)$ for each bag is expressed in Equation~\ref{eq:hem}. 

\begin{equation}
\label{eq:hem}
P(x_i) = \frac{\displaystyle\sum_{j=1}^{k} (k-j+1)p(x_{ij})}{\displaystyle\sum_{j=1}^{k} (k-j+1)}.
\end{equation}

The Weighted Evaluation Method assigns a higher weight $k$ to the first instance $x_{i1}$, that intuitively comes from the first saliency obtained from the Patch-SaliMap algorithm. Since the Patch-SaliMap in the first iteration achieves the highest value for the regions of the saliency map, this region has the major probability. The next saliency values are smaller than the first, so the algorithm assigns decreasing costs until the last saliency receives the weight equal to 1.

\section{Results}
\label{sec:results}

In this section, after describing our experimental setup (Section~\ref{sec:setup}), we report and discuss our empirical results on IP102~\citep{wu2019ip102}, a database for insect pest classification, and our Citrus Pest Benchmark (introduced in Section~\ref{sec:our-database}). In Section~\ref{sec:results-ip102}, we evaluate different CNNs on IP102 database. Next, in Section~\ref{sec:results-ourdatabase}, we explore %the key aspects of 
our proposal method on our benchmark, considering the best CNN evaluated on IP102.

\subsection{Experimental Setup}
\label{sec:setup}

We evaluated our experiments on five CNNs that are widely used in computer vision problems: Inception-v4~\cite{szegedy2017inception}, ResNet-50~\cite{he2016deep}, NasNet-A Mobile~\cite{zoph2017learning}, MobileNet-v2~\cite{sandler2018mobilenetv2}, and EfficientNet-B0~\cite{tan2019efficientnet}. We~chose these networks because they cover different common~features (and number of weights) presented in today's~CNNs.

We trained each CNN with Stochastic Gradient Descent with AdaDelta optimizer~\cite{zeiler2012adadelta}, batch size of up to 128, a learning rate of 0.1, weight decay of 0.0005, and cross-entropy function on top of the softmax output as a loss function. All CNNs are pre-trained on the ImageNet~\cite{krizhevsky2012imagenet} and then fine-tuned on the target database. We normalized the images, subtracting from the mean and dividing by the standard deviation, based on the ImageNet. For the experiments conducted on IP102, we resized all images to 224$\times$224 pixels.

We applied in training time an automatic data augmentation to our images. All of our experiments used a zoom range between $0.6$ and $1.4\times$, a rotation range between 0 and 360 degrees with values multiple of 15 degrees, vertical and horizontal reflection, and translation from 0 to 4 pixels along both axes.

To reduce overfitting, for IP102 database, we used dropout~\cite{srivastava2014dropout} between each of the EfficientNet-B0 modules (20\%), and after every depth-wise convolution (30\%). For Citrus Pest Benchmark, we also used dropout between each of the EfficientNet-B0 modules (20\%), after every depth-wise convolution (40\%) and before the final layer (30\%). % We also used L1 and L2 regularization techniques. % the 30\%.

We used the Gradient-weighted Class Activation Mapping (Grad-CAM) method~\cite{selvaraju2017grad} to extract the saliency maps.

Our models are trained on an NVIDIA RTX 5000 and an RTX 2080 Ti. We conducted all experiments using Keras/TensorFlow. Auxiliary code was developed using NumPy, Pandas and Scikit-Learn libraries. For the Grad-CAM\footnote{\url{https://github.com/jacobgil/keras-grad-cam}} and all CNNs (except for the EfficientNet\footnote{\url{https://github.com/qubvel/efficientnet/blob/master/efficientnet}}), we ran the experiments using the Keras implementation.

For every setup, we used five separate training sets to reduce the effects of randomness. The code and data are available at our Github repository\footnote{\url{https://github.com/edsonbollis/Weakly-Supervised-Learning-Citrus-Pest-Benchmark}}.

\subsection{Results for IP102}
\label{sec:results-ip102}

The IP102~\citep{wu2019ip102} database contains 102 classes and 75,222 images split into 45,095 training, 7,508 validation, and 22,619 test images for insect pest classification task. In addition, the database has a hierarchical structure and each sub-class is assigned with a super-class according to the type of damaged crops: field (e.g., rice, corn, wheat, beet, and alfalfa) and economic crop (e.g., mango, citrus, and vitis). All images were collected from the Internet.

We used this database to compare different CNN architectures to classify insect pests. The classification performance is evaluated using the standard metrics for this database, accuracy and F1-score.

Our results for IP102 are reported in Table~\ref{table:CNNs}. Not surprisingly, the EfficientNet (the state of the art in CNNs) reached the best classification performance (59.8\% of accuracy). However, we used its smallest version B0. That might indicate that the number reported does not represent the limit of classification performance achievable by the EfficientNet.

Regarding the number of weights (taking into account a mobile scenario), the MobileNet-v2, the smallest CNN in our experiments, reported 53.0\% of accuracy, an absolute difference of 6.8\% when compared to the EfficientNet performance.

\begin{table}[!htb]
\begin{center}
\small
\begin{tabular}{lcc}
\toprule
CNNs & Accuracy (\%) & Weights (M) \\
\midrule
Inception-v4 & 48.2 & 41.2\\
ResNet-50 & 54.2 & 23.6\\
NasNet-A Mob. & 53.4 & 4.4\\
\textbf{EfficientNet-B0} & \textbf{59.8} & 4.1\\
MobileNet-v2 & 53.0 & \textbf{2.3} \\

\bottomrule
\end{tabular}
\end{center}
\caption{Classification accuracy (in \%) results of different CNNs on the IP102 validation set. Here, we opted for evaluating on the validation set to \emph{not} optimize hyperparameters on the test set. Weights (in~M) mean the number of weights in millions of each CNN and the highlights in bold correspond to the best results.}
\label{table:CNNs}
\end{table}

For reference purposes, we show in Table~\ref{table:Reportes} the best results reported to date on the IP102 test set. The ResNet-50~\citep{wu2019ip102} result is the best outcome achieved by the dataset creators. They also reported statistics for the benchmark, which demonstrates that it is strongly unbalanced compared to other databases. The FR-ResNet~\citep{ren2019feature} approach changed the residual blocks internally, adding convolutions and reusing the initial features, since they hypothesized that the reuse of features from previous blocks improved the performance. They compared different types of convolutions in the blocks to the same number of parameters, since it is time consuming to test with many images in benchmarks as IP102. The DenseNet-121~\citep{xu2019xcloud} approach did not bring any information about how the authors reached the accuracy reported, neither how many times they trained the network nor if the value reported followed the database protocol. In addition, other metrics were not reported in their study, for instance, F1-score.

\begin{table}[!htb]
\setlength{\tabcolsep}{1.0mm}
\begin{center}
\small
\begin{tabular}{lccc}
\toprule
CNNs & Accuracy (\%) & F1-Score (\%) & Weights (M) \\
\midrule
ResNet-50~\citep{wu2019ip102} & 49.4 & 40.1 & 23.6\\
FR-ResNet~\citep{ren2019feature} & 55.2 & 54.8 & 30.8\\
DenseNet-121~\citep{xu2019xcloud} & \textbf{61.1} & N/A & 7.1\\
\textbf{EfficientNet-B0} & 60.7 & \textbf{59.6} & \textbf{4.1}\\
\bottomrule
\end{tabular}
\end{center}
\caption{Classification performance of different CNNs on the IP102 test set. Weights (in~M) mean the number of weights in millions of each CNN. N/A means that the value was not available from the original paper, The highlights in bold correspond to the best results.}
\label{table:Reportes}
\end{table}

\subsection{Results for Citrus Pest Benchmark}
\label{sec:results-ourdatabase}

In this section, we evaluate our method using EfficientNet-B0. As we show in Table~\ref{table:our-method}, we split the results into three parts, namely: 

\begin{itemize}

\item \textbf{Typical}: Since EfficientNet-B0 models require input images of 224$\times$224 pixels, we resize all images, distorting the aspect ratio to fit when needed. To highlight the mites in the convolutions, we also feed the network with the original image size of 1200$\times$1200 pixels.

\item \textbf{Baseline}: We first resize all images from 1200$\times$1200 to 897$\times$897 pixels. Next, we crop patches of size 299$\times$299 pixels and we manually select the ones with mites as positive samples. 

\item \textbf{Our Method} (detailed in Section~\ref{sec:our-method}): To make the comparisons fair, we extract patches of size 400$\times$400 from images of 1200$\times$1200 pixels, keeping the same ratio of the number of patches per image of baseline $\left(\displaystyle\frac{1200 \times 1200}{400 \times 400}=\frac{ 897 \times 897}{299 \times 299} = 9\right)$.

\end{itemize}

\begin{table}[!htb]
\setlength{\tabcolsep}{0.5mm}
\begin{center}
\small
\begin{tabular}{lc}
\toprule
EfficientNet-B0 & Accuracy (\%) \\
\midrule
\textbf{Typical}\\% Approach\\ Usual? Regular?
No patches, 224$\times$224 pixels	   & 75.9 \\
No patches, 1200$\times$1200 pixels     & 81.2 \\
\cmidrule(lr){1-2}
\textbf{Baseline} \\ 
Manually-annotated patches, 299$\times$299 pixels  & 86.0 \\ 
\cmidrule(lr){1-2}
\textbf{Our Method} \\
Automatically-generated patches, 400$\times$400 pixels & \textbf{91.8} \\ 
\bottomrule
\end{tabular}
\end{center}
\caption{Classification accuracy (in \%) results on the Citrus Pest Benchmark validation set. Here, we opted for evaluating on the validation set to \emph{not} optimize hyperparameters on the test set. We split the results into three parts, namely: Typical, Baseline, and Our Method. The value highlighted in bold corresponds to the best result.}
\label{table:our-method}
\end{table}

The ``overall picture'' from Table~\ref{table:our-method} can be summarized as follows. Our Method surpasses the classification performance over all schemes. The comparison between Typical results shows that --- as usually observed for image classification~\cite{valle2020ddd,Mendoza2020} --- high-resolution images lead to better performance. Baseline scenario (manually-annotated patches) shows promising results, however, annotating patches is a tedious task, time-consuming, and error-prone. In comparison to Typical result (no patches, 1200$\times$1200 pixels), Baseline --- even using patches of size 299$\times$299 pixels --- significantly increases the classification performance, indicating that the model can benefit from patch representations. Comparing Our Method to Baseline (automatically-generated patches vs. manually-annotated patches), we observe an increase from 86.0\% to 91.8\%, an absolute improvement of~5.8\%.

Our best model in the test set (we restricted ourselves to perform experiments in validation set) achieved an accuracy of~92.1\%.

\begin{figure}[!htb]
\captionsetup[subfloat]{farskip=2pt,captionskip=2pt}
\centering
\subfloat[][Input image]{\includegraphics[height=0.23\textwidth]{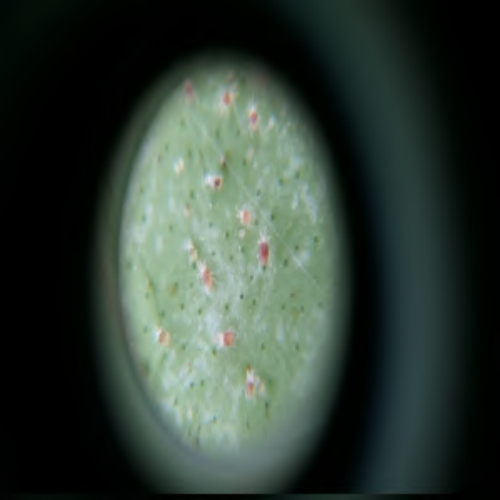}\label{pics:Grad-CAM-a}}\hspace{.02cm}
\subfloat[][Saliency map]{\includegraphics[height=0.23\textwidth]{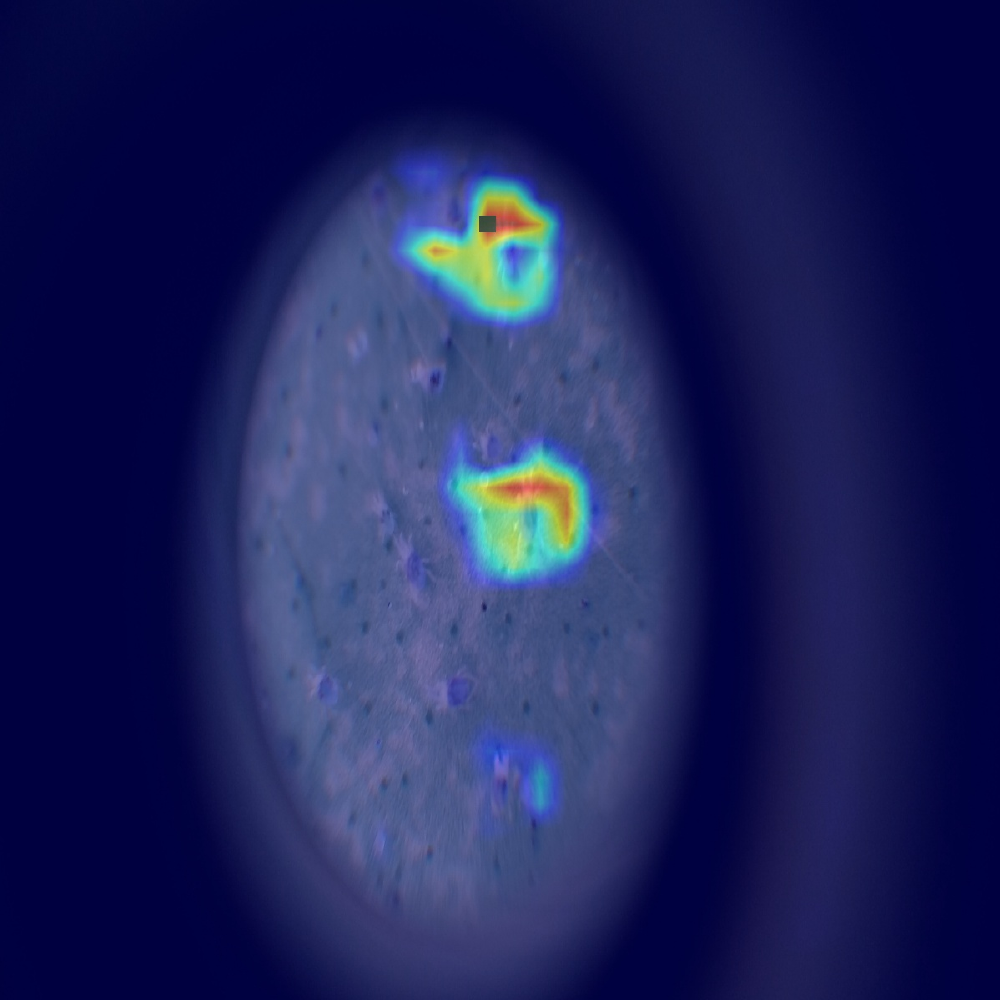}\label{pics:Grad-CAM-b}}\hspace{.02cm}
\subfloat[][Patch 1]{\includegraphics[height=0.14\textwidth]{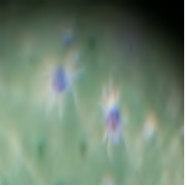}\label{pics:Grad-CAM-c}}\hspace{.02cm}
\subfloat[][Patch 2]{\includegraphics[height=0.14\textwidth]{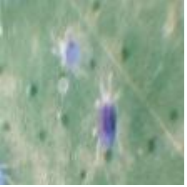}\label{pics:Grad-CAM-d}}\hspace{.02cm}
\subfloat[][Patch 3]{\includegraphics[height=0.14\textwidth]{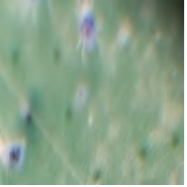}\label{pics:Grad-CAM-e}}\hspace{.02cm}
\subfloat[][Patch 4]{\includegraphics[height=0.14\textwidth]{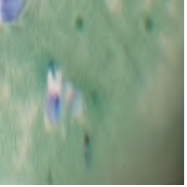}\label{pics:Grad-CAM-f}}\hspace{.02cm}
\subfloat[][Patch 5]{\includegraphics[height=0.14\textwidth]{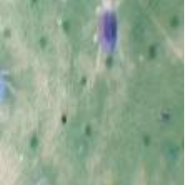}\label{pics:Grad-CAM-g}}\hspace{.02cm}
\caption{Automatically-generated patches guided by the saliency map.}
\label{pics:Grad-CAM}
\end{figure}

For illustration, we show in Figure~\ref{pics:Grad-CAM} the automatically-generated patches guided by the saliency map. The patches are ranked according to the highest activation (from Figure~\ref{pics:Grad-CAM-c} to~\ref{pics:Grad-CAM-g}. The generated patches highlight the positive impact of our method.

\begin{comment}
\begin{figure}[!htb]
\captionsetup[subfloat]{farskip=2pt,captionskip=2pt}
\centering
\subfloat[][Original image]{\includegraphics[height=0.23\textwidth]{./images/20180302_080753.pdf}\label{pics:Grad-2-a}}\hspace{.02cm}
\subfloat[][Saliency Map]{\includegraphics[height=0.23\textwidth]{./images/cgradcam_20180302_080753.pdf}\label{pics:Grad-2-b}}\hspace{.02cm}
\subfloat[][Cut 1]{\includegraphics[height=0.13\textwidth]{./images/ccutted_0_20180302_080753.pdf}\label{pics:Grad-2-c}}\hspace{.02cm}
\subfloat[][Cut 2]{\includegraphics[height=0.13\textwidth]{./images/ccutted_1_20180302_080753.pdf}\label{pics:Grad-2-d}}\hspace{.02cm}
\subfloat[][Cut 3]{\includegraphics[height=0.13\textwidth]{./images/ccutted_2_20180302_080753.pdf}\label{pics:Grad-2-e}}\hspace{.02cm}
\subfloat[][Cut 4]{\includegraphics[height=0.13\textwidth]{./images/ccutted_3_20180302_080753.pdf}\label{pics:Grad-2-f}}\hspace{.02cm}
\subfloat[][Cut 5]{\includegraphics[height=0.13\textwidth]{./images/ccutted_4_20180302_080753.pdf}\label{pics:Grad-2-g}}\hspace{.02cm}
\caption{Cut images from Grad-CAM using the Instance Model. The black square is the heist value produced by the Grad-CAM.}
\label{pics:Grad-2}
\end{figure}
\end{comment}

\section{Conclusions and Future Work}
\label{sec:conclusions}

In this work, we presented a new weakly supervised Multi-Instance Learning (MIL) process to classify tiny regions of interest (ROIs), a Selection Strategy Based on Saliency Maps (Patch-SaliMap), a Weighted Evaluation Method, as well as a novel database for agriculture called Citrus Pest Benchmark (CPB).

The CPB is the first database containing images acquired via mobile devices from citrus plants for pest recognition. A number of different mite species, typically invisible to the naked eye, may affect citrus leaves and fruits. The benchmark is a valuable resource for the automation of Integrated Pest Management (IPM) tasks in agriculture and for the evaluation of new classification algorithms.

From our experiments, we observed that our classification method was able to achieve superior results when compared to other approaches on the IP102 dataset. In addition, we discussed the effectiveness of our method on the CPB dataset, surpassing two different experimental scenarios. The weakly supervised multi-instance learning approach demonstrated to be effective in identifying patches of interest. The strategy for selecting the multiple patches reduced the probability of losing relevant regions, consequently improving our classification results. Overall, we believe that our method has great potential to help inspectors to classify pests and diseases through magnifying glasses and mobile devices directly in the field.

As directions for future work, we plan to further analyze the EfficientNet attention modules, so they can better operate in small areas of the images. This could reduce those patches without the occurrence of mites produced by the Patch-SaliMap. Moreover, we will investigate how small differences among the mite species would affect the multi-class task. Finally, we intend to deploy our CNN-based learning process on mobile devices.

\section*{Acknowledgments}

E. Bollis is partially funded by CAPES (88882.329130/2019-01). H. Pedrini is partially funded by FAPESP (2014/12236-1, 2017/12646-3) and CNPq (309330/2018-1). S. Avila is partially funded by FAPESP (2013/08293-7, 2017/16246-0) and Google Research Awards for Latin America 2019. RECOD Lab. is partially supported by diverse projects and grants from FAPESP, CNPq, and CAPES. We gratefully acknowledge the donation of GPUs by NVIDIA Corporation.

{\small
\bibliographystyle{abbrvnat}
\bibliography{references}
}

\end{document}